\documentclass[runningheads]{llncs}
\usepackage[T1]{fontenc}
\usepackage{graphicx}

\begin{document}
\title{When to Make a Decision: Task-Sufficient Active Perception for Mobile Manipulator Robots}
\titlerunning{When to Make a Decision}

\author{Mariana da Silva Barros\and
Matthias Kerzel\and
Cornelius Weber\and
Stefan Wermter}

\authorrunning{M. da Silva Barros et al.}

\institute{Knowledge Technology, Department of Informatics, University of Hamburg, Hamburg, Germany\\
\email{\{mariana.barros, matthias.kerzel, cornelius.weber, stefan.wermter\}@uni-hamburg.de}}

\maketitle              
\begin{abstract}
A robotic agent operating in an unknown environment must gather sufficient task-relevant information before acting. However, determining when the available information is sufficient remains a challenge. We propose a conceptual hybrid neuro-symbolic framework for task-sufficient active perception in unknown environments. After interpreting a requested task, the agent generates an initial plan and iteratively reevaluates it as new information becomes available. At each iteration, it can acquire a more informative viewpoint, inspect or manipulate candidate objects, ask the user for clarification, or perform the task. Neural perception modules update an object-centric belief state, while a symbolic decision layer determines whether the remaining uncertainty justifies further information gathering. The framework will be implemented on a mobile manipulator comprising a humanoid torso and a mobile base, with complementary evaluation in simulation. The proposed approach aims to enable robots to determine not only how to gather additional information, but also when further perception is no longer necessary.

\keywords{Active Perception \and Mobile Manipulation \and Neuro-symbolic Robotics \and Task Planning \and Decision-making Under Uncertainty}
\end{abstract}

\section{Introduction and Motivation}
\label{sec:Intro}

Advances in robotics and artificial intelligence are enabling robots to perform an increasingly broad range of assistive tasks. However, when placed in an unknown and dynamic environment before performing a task, the robotic agent must perceive and gather sufficient information about the surroundings, which is often challenging for robots to do \cite{Morilla-Cabello_2024}.

\begin{figure}
\centering
\includegraphics[width=0.6\textwidth]{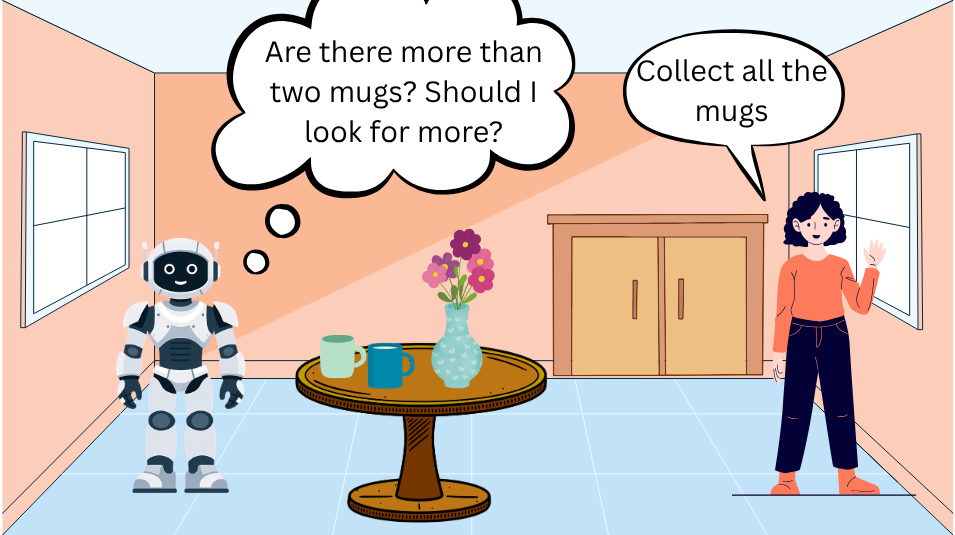}
\caption{Illustrative scenario for a task under occlusion and insufficient information.} \label{fig_ex_mugs}
\end{figure}

Consider a robot operating in a previously unseen environment and receiving a task such as "collect all the mugs in the room". The example is illustrated in Figure \ref{fig_ex_mugs}. The robot has sensors to collect information about its surroundings, but in its current position, it can only observe part of the environment. After making an initial plan based on the current situation, at each iteration, the robot can explore the environment, inspect or manipulate objects, or ask the user for clarification. Only when the robot has sufficient relevant knowledge can it perform the requested task.

Depending on different factors such as the environment, the required task, user preferences, or how much information the robot has already gathered, there may be several choices for each situation. In the above mug scenario, the robot can see two mugs on the table, but some viewpoints are obstructed by the vase from its current position. Therefore, it could explore further the surroundings or seek a better point of view (on the other side of the table). Another option would be to open the closet and see if there are mugs inside. Once it finds a mug in the room, it can either explore to see if there are other mugs, pick it up and deliver it to the user, or ask the user if there are more mugs present in the room. If it acts too soon, it may miss unseen objects and cause the task to fail. However, if the robot explores for too long, it may only get redundant or unnecessary information and waste time.

This scenario motivates the need for a framework that focuses on the task-relevant view of active perception. The main goal is for the robotic agent to acquire enough task-relevant knowledge to act safely and effectively. Or, in other words, for a robot to determine when its current knowledge is sufficient for task execution in an unknown environment. If the robot still does not have enough information to perform the task, it must move its base, head position, or hand position to collect more necessary data. Therefore, this work proposes investigating how a mobile robot with a manipulator can determine when its current knowledge is relevant and sufficient for task execution in an unknown environment. The conceptual work proposed includes modules that will perform task interpretation and planning, active perception, object inspection, and interaction with the user.

\section{Related Work}

In recent years, various approaches have been developed in different areas related to the execution of robot tasks, including perception, object detection, navigation, and task planning. In real-world environments, where object occlusion is particularly common (due to the environment itself or other objects), static perception is not sufficient for the robot to observe the environment completely. Therefore, in such scenarios, it is essential for robotic agents to perform active perception, where a robot actively moves its body and sensors to acquire relevant knowledge from the environment \cite{Active_perception_1988}. This ability allows the robotic agent to determine what it wants to perceive and a way to achieve this \cite{Bajcsy_2018}. 

Scene understanding is crucial for task execution in different situations, especially in unknown and dynamic environments. For robotic agents, it is vital to ensure robustness in grasping movements in cluttered scenes \cite{Jauhri_2024}. Active vision can also support navigation by directing the robot toward viewpoints expected to improve localization accuracy \cite{Barkalas_2024}. For mobile manipulation, active-perceptive methods can jointly optimize viewpoint selection, base motion, and grasp feasibility \cite{Jauhri_2024}. 

Recent advances in computer vision and machine learning have enabled different approaches in active perception. A perception model \cite{Morilla-Cabello_2024} is proposed that improves the estimation of information gain using perceptual factors to quantify the aggregated influence of the environment on candidate viewpoints. However, this work uses only a mobile robot, not including manipulation or interaction with objects or the environment. Jauhri et al. \cite{Jauhri_2024} propose an active perception pipeline for mobile manipulators that balances exploration (maximizing visual information gain) and exploitation of task-specific information (such as grasp executability for mobile grasping). However, the robot can get stuck in deadlocks while trying to switch targets, leading to oscillating motions. Another limitation is the limited observability in occluded scenes, making the information gain computation difficult. Another approach consists of an active vision-based path planning method that, in addition to robust path planning and obstacle avoidance, guides the robot to areas with the most robust visual features to further refine the localization error \cite{Barkalas_2024}. However, this approach generates paths that are longer than necessary and is also constrained by an empirically selected static threshold used to assess the probability of tracking failure in localization.

Despite the various approaches that implement perception for mobile robots, they often separate planning, perception, exploration, and interaction into distinct modules, which are not motivated by the relevance and sufficiency of the underlying knowledge. As a result, either the robot may gather irrelevant information (not necessary for the task) or may fail to gather sufficient data or correctly interpret the task. In addition, there is a gap in the efficient use of all the robot's capabilities (motion, manipulation, and human interaction) to support task execution.

\section{Proposed Framework}

\begin{figure}
\includegraphics[width=\textwidth]{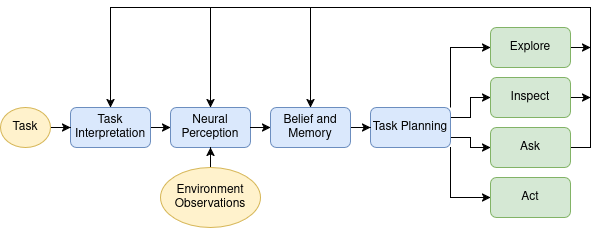}
\caption{Proposed uncertainty-aware framework and decision loop.} \label{fig_archi}
\end{figure}

In this work, we propose a hybrid neuro-symbolic framework that enables an embodied agent to perceive an unknown environment and to treat exploration, inspection, clarification, and task execution as alternative actions within a unified uncertainty-aware control loop. The overview of the proposed architecture is presented in Figure \ref{fig_archi}, and the individual modules are described below.

The proposed framework builds on the modular design principles of the ELMiRA framework \cite{Elmira_2024}, which demonstrates modular neural Human-Robot Interaction (HRI) for robot action. Whereas ELMiRA maps language-mediated interaction to robot actions, the proposed system extends this architecture to persistent decision-making under partial observability. It introduces an object-centric belief state, explicit uncertainty estimation, active information-gathering actions, and a task-sufficiency criterion that determines when the robot should stop gathering information and act.

\subsection{Task Interpretation}

The process begins when the robot enters an unknown environment and receives a task instruction. This module will use a large language model (LLM) to interpret natural-language instructions and translate them into structured task representations.

In a simple real-world scenario considered in this paper, possible tasks are: object search, object retrieval, and navigation towards a specific part of the environment. Once the robot receives a requested task, it must interpret it in a way that enables it to act and also to explain its actions during the execution of the task.

\subsection{Neural Perception}

The perception layer will use neural models for object detection, semantic segmentation, and vision-language grounding. For object detection, we will compare Convolutional Neural Networks (CNNs) and transformer-based models, and for language grounding, a Vision-Language Model (VLM) will be used. These models provide candidate object labels, visual embeddings, spatial estimates, and confidence scores that are incorporated into a belief state, which is updated through movement, inspection, and clarification. 

\subsection{Belief and Memory}

Once the robot learns the task to be done and has perceived the scene from its current point of view, it will update its memory module, which will be implemented as an object-centric semantic memory \cite{liu2023active}. Each remembered object candidate will be represented by its estimated location, semantic category, detection confidence, observation history, and learned visual-language embeddings. This will enable the robot to reason not only about where the objects are, but also about whether their identity and task relevance are sufficiently certain.

\subsection{Decision Layer}

After processing the sensory input and performing the perception, the robotic agent must make a decision regarding its next step, which is the core of the system architecture. The task-sufficiency criterion will combine the uncertainty of the neural perception output with spatial and task-level uncertainty to decide the action that will follow. For example, low detector confidence may trigger further inspection, while multiple visually similar candidates or ambiguous language grounding may require further clarification. This decision can be compared to balancing between exploration and exploitation in Reinforcement Learning \cite{Zangirolami_2024}. The possible actions for the robot are described below.

\begin{itemize}

    \item \textbf{Active Perception:} When the robot is in an unknown relevant area, it may further explore the environment surroundings. Active perception will enable the agent to seek a more suitable point of view to locate objects or find a better path. This module will be implemented using viewpoint selection based on expected reduction in semantic uncertainty \cite{Burusa_2024}, and this can be performed using a navigation approach, which is formulated as Next-Best-View \cite{NBV_2016}.\\

    \item \textbf{Inspection:} Another action that the robot can perform is to inspect and manipulate objects present in the environment, in the case that the candidate object needs better observation. For example, if the task is to "bring me the empty boxes", one efficient way to do that would be to lift the objects and use the weight information \cite{Kerzel_2019}. In addition, the agent can interact with the environment to gather more information, for example, opening cabinets or drawers.\\

    \item \textbf{Clarification:} The agent can face problems of uncertainty, ambiguity, or vagueness \cite{Sun_2026}, and, in this case, it must ask the user for confirmation or ask questions about the goal task. The HRI module for this communication will be based on the ELMiRA framework \cite{Elmira_2024}.\\

    \item \textbf{Task Execution:} Once the robot's confidence exceeds a defined threshold, it will execute the required task.
    
\end{itemize}

Given the perception results about the environment and objects surrounding the embodied agent, it must choose one of the possible actions. Decision-making will be carried out using a behavior tree approach, such as the one used in ELMiRA \cite{Elmira_2024}, that will be adapted for our framework, with the use of neural transforms for sensing and a symbolic behavior tree for planning and actions.

\section{Experiment Setup and Evaluation}

We will develop and evaluate the system both in simulation and in a real-world environment, which will enable the training of the model in a safe environment, but also the deployment and testing in real-world situations. Sim-to-real transfer will be performed using domain adaptation techniques \cite{Gaede_2024}.

\subsection{Implementation}

The simulation environment will be implemented in the NVIDIA Isaac Sim\footnote{https://developer.nvidia.com/isaac/sim}. This framework enables the creation of realistic environments with accurate physical properties, which not only allows for model training but also for reproducible testing and safe debugging. In order to train and test the system, different indoor environments will be simulated, including realistic object arrangements, varying degrees of occlusion, perceptually similar objects, and ambiguous task instructions.

For the physical implementation, the target platform will be an agent built as a combination of a humanoid torso on a mobile platform. This embodied agent combines the mobility of a wheeled robot (which enables easier navigation and environmental exploration) with the reachability and manipulation ability of a bimanual robot. In addition, the humanoid features of the agent make it possible to acquire observations from a human-like sensor configuration, with cameras simulating the vision and the ability to turn the head in different directions to acquire observations from different viewpoints.

\subsection{Planned Evaluation}

We will evaluate the systems in different scenarios that can be implemented in a laboratory and office environment, and can also be reproduced. The proposed technique will be evaluated with common tasks requested of the robot in different scenarios and levels of complexity, simulating real situations.

The goal of the described experiments is to assess different aspects of the system, such as the task success rate, the path and action plan created by the robot, and the number of steps necessary to reach a goal. In addition to task success and exploration efficiency, the evaluation will analyze how the uncertainty of neural perception changes after exploration, inspection, or clarification, and whether the robot avoids acting on low-confidence or ambiguous predictions.

\section{Expected Contributions}

The expected contribution is a hybrid neuro-symbolic architecture for task-sufficient active perception in mobile manipulation. Building on the modular HRI structure of ELMiRA, the framework introduces an object-centric belief state and a task-sufficiency decision layer that selects between exploration, object inspection, user clarification, and task execution. Unlike approaches that optimize a single perception mechanism, the proposed system compares heterogeneous information-gathering actions according to their expected reduction in task-relevant uncertainty, cost, and risk. The framework will be evaluated in reproducible simulated and physical scenarios that involve occlusion, perceptual ambiguity, and incomplete task specifications.

In addition, the core idea of this work is to develop an architecture that is adaptable across different environments and platforms and applicable to real-world scenarios. The system will be evaluated with common reproducible tasks in small, unknown environments.

\section{Conclusion}

This work proposes a task-sufficient approach to active perception for mobile manipulation. The planned architecture prioritizes reasoning about whether its current knowledge is sufficient for the task at hand, rather than pursuing a complete scene understanding. Integration of embodied memory, uncertainty estimation, active exploration, and human clarification intends to improve robustness in unknown environments. Furthermore, the proposed implementation will provide support for testing in real-world scenarios and will enable the study of how mobile manipulators decide when to further perceive and when to act.

\begin{credits}
\subsubsection{\ackname} The authors gratefully acknowledge support from the German Research Foundation DFG under project LUMO (No 551629603) and funding from Horizon Europe under the MSCA grant agreements No 101226624 (GREET), No 101168792 (SWEET), and No 101072488 (TRAIL).

\subsubsection{\discintname} The authors have no competing interests to declare that are relevant to the content of this article.
\end{credits}

\bibliographystyle{splncs04}
\bibliography{mybib}
\end{document}